%% file: main.tex
\documentclass[11pt]{article}

\usepackage[final]{acl}
\usepackage{graphicx}
\usepackage{subcaption}
\usepackage{multirow}
\usepackage{amsmath}
\usepackage{amssymb}
\usepackage{times}
\DeclareSymbolFont{operators}{OT1}{ptm}{m}{n}
\SetSymbolFont{operators}{bold}{OT1}{ptm}{b}{n}
\usepackage{latexsym}
\usepackage[T1]{fontenc}
\usepackage[utf8]{inputenc}
\usepackage{microtype}
\usepackage{inconsolata}
\usepackage{booktabs}
\usepackage{tcolorbox}
\tcbuselibrary{breakable}
\usepackage{enumitem}

\title{SnowballCost: A Systematic Measurement of the Impact of Ultra-Long Responses on Large Language Model Inference Services}

\author{Xijia Feng$^{1}$ \quad Han Qiu$^{2}$\thanks{Corresponding author.} \quad Jianshuo Dong$^{2}$ \quad Biplab Sikdar$^{1}$ \\
  $^{1}$National University of Singapore \quad $^{2}$Tsinghua University \\
  \texttt{xijia.feng@u.nus.edu} \quad \texttt{qiuhan@tsinghua.edu.cn} \\
  \texttt{dongjs23@mails.tsinghua.edu.cn} \quad \texttt{bsikdar@nus.edu.sg}}

\begin{document}
\maketitle

\begin{abstract}
Modern Large Language Models (LLMs) support ever-larger context windows and output budgets. Under autoregressive decoding, the cost of generating a response grows faster than its length, yet the cost that ultra-long responses impose on LLM inference services remains underexplored.
We present \textsc{SnowballCost}, a framework that measures this cost across serving engines, models, and GPU configurations.
From the measurements we derive an inference cost model and calibrate it. Its coefficients are set by the model architecture, the GPU specifications, and the serving engine, so one calibration carries over to other deployments. The cost model also gives two threshold lengths, one at which a linear cost estimate fails and one at which a request stops being profitable.
We find that (1) even optimized engines such as vLLM and SGLang exhibit super-linear growth in inference time at extreme response lengths, up to 1M tokens.
(2) When served alongside requests with ultra-long responses, normal requests are only slightly affected, and their SLO attainment barely changes. The requests with ultra-long responses themselves take longer to generate each token as their output grows, and a few of them occupy the provider's GPUs long after the normal traffic has finished.
(3) An ultra-long response can become unprofitable before it completes.
Our framework, request traces, and analysis scripts are available at \url{https://github.com/fengxijia/SnowballCost}.
\end{abstract}

\section{Introduction}
\begin{figure}[t]
\centering
\includegraphics[width=0.99\columnwidth]{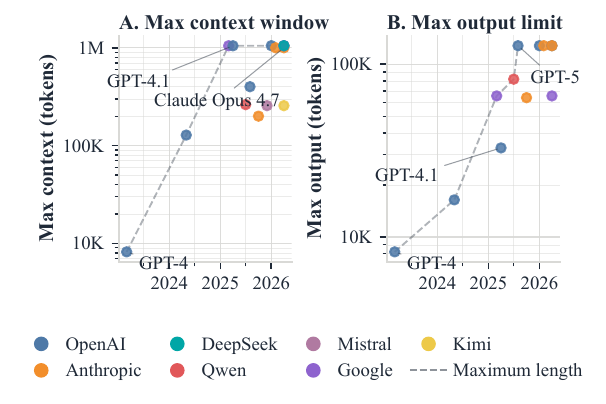}
\vspace{-1em}
\caption{Maximum context window and maximum output length of frontier LLMs by release date. Since 2023, the largest context window and the largest output limit have each grown by more than an order of magnitude.}
\vspace{-1.5em}
\label{fig:trend}
\end{figure}

For service providers, inference now consumes more of the compute spent on Large Language Models (LLMs) than training does~\cite{deloitte2025compute, fernandez-etal-2025-energy}. With frontier LLMs now supporting million-token or even longer context windows~\citep{
deepseekai2026deepseekv4,
googledeepmind2026gemini31pro,
openai2026gpt55,
anthropic2026claude_context,
yang2025qwen25_1m},
responses can grow correspondingly long (Figure~\ref{fig:trend}). Under autoregressive decoding, the computation needed for a response snowballs with its length. Cost estimates for inference services seldom account for this growth in \textit{ultra-long responses}.
The cost of ultra-long responses is a \textit{long-output} problem, which differs from the \textit{long-context} problem~\cite{ruler,beyondlimits}. Long-context research studies the cost of \textbf{reading} long inputs during prefill, whereas we study the cost of \textbf{generating} long outputs during decode. Prior work quantifies the inference energy~\cite{samsi2023words, poddar-etal-2025-towards} or demonstrates prompts that trigger long outputs~\cite{engorgio}, but the system-level impact of ultra-long responses on providers and users remains unexamined.

This paper asks: \emph{what implications do ultra-long responses have for LLM service providers and users?} To answer this question, we develop \textsc{SnowballCost}, a systematic \textbf{measurement framework} that comprises a request generator, the inference engines under test, and a metrics monitor. 
We derive a \textbf{cost model} from the memory traffic of each decode step and calibrate it on these measurements. After calibration on two output lengths, the cost model predicts the decode time of a 1M-token output of Qwen2.5-7B-1M on vLLM within $4.3\%$. The same calibration applies to other models, GPU types and device counts (Section~\ref{sec:estimator}). The cost model also gives two \textbf{threshold lengths} for providers.
Our measurements and model yield three findings:
(1) single-request decode time grows super-linearly with output length, and for a 1M-token output of Qwen2.5-7B-1M it is $1.80\times$ a linear extrapolation from 131K on vLLM and $1.67\times$ on SGLang (Section~\ref{sec:single_request});
(2) shared serving mainly affects the requests with ultra-long responses and the provider, and affects normal requests only slightly (Section~\ref{sec:mixed}). The time per output token of each request with an ultra-long response grows with its output length and with the number of these requests on the node. When $1\%$ of the requests have ultra-long responses, these requests keep the provider's node busy about $100\times$ as long as the normal traffic they arrive with;
(3) revenue grows linearly with output tokens while decode time grows super-linearly, so a request becomes unprofitable beyond a threshold length that our cost model gives in closed form (Section~\ref{sec:cost_model}).

\begin{figure*}[!t]
\centering
\includegraphics[width=0.99\textwidth]{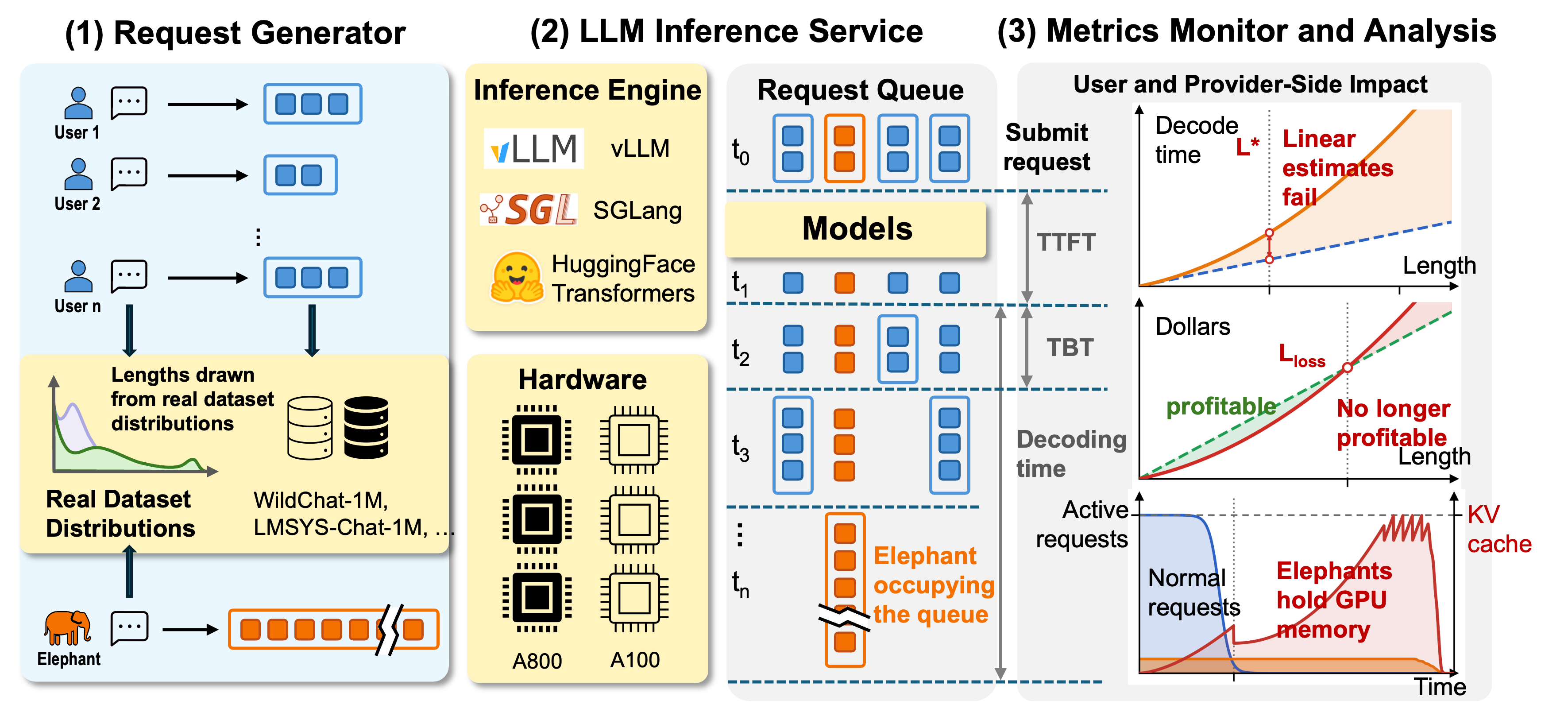}
\caption{Overview of the \textsc{SnowballCost} measurement framework. A \textbf{request generator} draws prompts from real-world datasets and mixes normal traffic with a set fraction of requests that produce ultra-long responses. We call these requests \emph{elephants}. Requests are dispatched to \textbf{inference engines} running on GPU platforms. A \textbf{metrics monitor} records client-side latency together with server-side metrics such as KV-cache usage, so per-request latency and server load are analyzed on one timeline.}
\vspace{-2mm}
\label{fig:pipeline}
\end{figure*}

\section{Background and Related Work}
\label{sec:prelim}

\subsection{LLM Inference as a Service}
\label{sec:prelim_llmaas}

Unlike classic DNN inference, an autoregressive LLM request persists for many decode steps while its Key-Value (KV) cache grows. Early model-serving systems such as TensorFlow Serving, Nexus, and InferLine~\cite{tfserving,nexus,inferline} target generic DNN inference with single forward passes.

Modern LLM serving systems optimize this iterative process through specialized batching, scheduling, and memory management: iteration-level scheduling~\cite{orca}, paged KV-cache memory~\cite{vllm}, a structured front-end with KV reuse and continuous batching~\cite{sglang}, and improved tradeoffs between prefill and decode, KV-cache utilization and multi-model serving~\cite{sarathi_serve,mooncake,muxserve,fastserve}, on top of common model interfaces and kernel-level optimization~\cite{wolf2020transformers,fastertransformer}. DistServe~\cite{zhong2024distserve} disaggregates prefill and decode onto separate instances and fits a per-phase latency model to choose the placement and parallelism that serve the most requests within their SLOs; its decode term grows with the total context length of the batch, and it is fitted in the normal regime of hundreds to a few thousand tokens. Vidur~\cite{vidur} predicts operator runtimes with random forests, is validated on requests of at most 4K tokens, and must be re-profiled for every deployment. Sheng et al.~\cite{sheng2024fairness} fit a request-level quadratic cost on profiles with outputs of up to about 200 tokens and use it as a scheduling weight. Per-step bandwidth expressions~\cite{shazeer2019mqa, wang2026latencydos} give the time of one decode step. We extend these models to ultra-long outputs. We sum the per-step cost into a request-level closed form whose coefficients have physical units, calibrate it at 128K to 1M output tokens, and derive the output length at which the quadratic term exceeds the linear one.

\looseness=-1 Although these systems make LLM-as-a-service practical, they are mostly evaluated under short or moderately long workloads. In contrast, this paper studies shared serving, where normal requests and requests with ultra-long responses run together on the same server. We measure how a small fraction of requests with ultra-long responses affects it through decode-time residency, KV-cache pressure, and throughput degradation. Related work on inference energy and carbon footprint~\cite{llmcarbon,samsi2023words,dynamollm,sweetspot2026} quantifies aggregate or normal-regime inference cost, but does not isolate the serving-side impact of ultra-long generated outputs.

\subsection{Long Outputs in LLM Services}
\label{sec:prelim_long_output}

Commercial APIs already expose large generation budgets, including 128K-token or larger limits~\cite{openai_gpt51_docs,anthropic_models_docs}. Ultra-long responses come from three sources.

\noindent(1) \textit{Reasoning traces:}
Reasoning models may run tens of thousands of decode steps before producing a final answer. DeepSeek-R1-0528 reports average token usage of 23K on AIME 2025~\cite{deepseek_r1_0528}. OpenAI's reasoning API documentation notes that reasoning tokens count against the output budget and may exhaust \texttt{max\_output\_tokens} before any visible answer is returned~\cite{openai_reasoning_api}. Kimi k1.5 also highlights the need to control long reasoning rollouts to avoid monopolizing compute~\cite{kimi_k15}.

\noindent(2) \textit{Long-form generation:}
Applications such as solving hard problems and drafting documents can naturally require a large number of output tokens. LongWriter demonstrates coherent generation beyond 20K words via task decomposition~\cite{longwriter}. LongGenBench evaluates long-form instruction following at 16K and 32K tokens~\cite{longgenbench}. Recent work benchmarks long procedural generation~\cite{longproc} and calls for shifting attention from long-input processing to long-output generation~\cite{shifting_input_output}.

\looseness=-1 \noindent(3) \textit{Induced long-output generation:}
Output length can also be deliberately inflated by prompt design or generation controls. Engorgio crafts prompts that drive models into abnormally long generations~\cite{engorgio}. AutoDoS automates black-box prompt construction and reports large latency amplification~\cite{crabs}. Verbose-image prompts cause vision-language models to generate much longer descriptions~\cite{verbose_image}.

\section{The SnowballCost Measurement Framework}
\label{sec:methodology}
We define requests whose responses exceed a long-output threshold $L_{\mathsf{thr}}$ as \textbf{elephant requests}, and all other requests as \textbf{normal requests}.
To systematically quantify the impact of ultra-long responses on LLM inference services, we propose \textsc{SnowballCost} as illustrated in Figure~\ref{fig:pipeline}. Our framework consists of three core components: \textit{i)} a \textit{request generator} that emulates real-world request patterns of LLM inference services and mixes normal traffic with elephant requests at a set elephant fraction; \textit{ii)} the \textit{inference engines} under test; and \textit{iii)} a \textit{metrics monitor} that records metrics on both the provider side and the user side.
Ultra-long responses arise from diverse workloads, such as long reasoning, agentic loops and long-form generation, and their serving cost depends on the number of tokens generated, whatever produced them. We therefore set the output length of each elephant request directly.

\subsection{Request Generator and User Modeling}
\label{sec:generator}
Normal requests take their prompts and output lengths from real conversation datasets. Each request is an elephant request with probability $p_{\mathrm{ele}}$, which we call the elephant fraction. An elephant request generates a response of length $\eta L_{\max}$, a fraction $\eta$ of the context window $L_{\max}$ that lies above the long-output threshold $L_{\mathsf{thr}}$. In our simulations, requests follow two arrival patterns, a Poisson process and a burst in which they all arrive at once, which reflect the Poisson arrivals used to evaluate serving systems~\cite{vllm, zhong2024distserve} and the bursty traffic of production traces~\cite{wang2024burstgpt}.

\subsection{Inference Engines}
\label{sec:inference_engine}
We evaluate two production-grade open-source serving engines as our primary backends, and include HuggingFace Transformers as a controlled reference that exposes the quadratic cost.

\noindent\textbf{vLLM}~\cite{vllm} employs \textit{PagedAttention}, which stores the Key-Value (KV) cache in non-contiguous memory blocks to reduce fragmentation, and it uses continuous (iteration-level) batching so that elephant and normal requests are processed concurrently in dynamic batches. We run all mixed-workload experiments on vLLM.

\noindent\textbf{SGLang}~\cite{sglang} builds on similar kernel optimizations but introduces RadixAttention for cross-request KV reuse and finer-grained scheduling. Comparing SGLang with vLLM lets us observe how different system-level designs handle memory pressure under ultra-long responses.

\noindent\textbf{HuggingFace Transformers (reference baseline).}
HuggingFace Transformers~\cite{wolf2020transformers} is \emph{not} a production backend; we use it only as a controlled reference where the quadratic attention cost appears in its raw form, before the kernel and memory optimizations of production engines attenuate it: Eager mode exposes the raw $\mathcal{O}(l^2)$ growth, and Flash-Attention 2 (FA2)~\cite{dao2024flashattention2} reduces traffic to high-bandwidth memory (HBM) without reducing the arithmetic, so quadratic growth reappears at ultra-long lengths.

\subsection{Metrics Monitor}
\label{sec:monitor}
To quantify how ultra-long responses affect an inference service, we collect server- and client-side metrics and align them by request ID. Client-side latency can then be attributed to server-side queueing, decode, or memory pressure.

\noindent\textbf{Server-side metrics.}
\textit{Throughput} ($TPS$)~\cite{orca,vllm}: output tokens per second.
\textit{SLO attainment}~\cite{zhong2024distserve,sarathi_serve}: fraction of requests meeting their class's service-level objective (SLO), the latency budget set for that class.
\textit{Lifecycle timings}~\cite{splitwise,zhong2024distserve}: per-request prefill and decode times.
\textit{Concurrency state}~\cite{orca}: running and waiting requests.
\textit{KV-cache utilization}~\cite{vllm,mooncake}: paged-cache occupancy.

\noindent\textbf{Client-side metrics.}
\textit{TTFT}~\cite{sarathi_serve, vidur, thrott}: time to the first response token. \textit{TBT}~\cite{sarathi_serve, vidur, splitwise}: average inter-token latency. \textit{E2E}~\cite{splitwise, clone, thrott}: submission to the last token. \textit{Inference time}: TTFT plus decode time; with the short prompts sampled from real chat data (median length of a few hundred tokens) it is decode-dominated, and we report it for measured single-request curves while the cost model targets decode time.

\section{Snowball Cost and Threshold Length Modeling}
\label{sec:modeling}
In this section, we model the inference cost of an elephant request. We fit the coefficients of the cost model by regression on measurements from our SnowballCost framework (Section~\ref{sec:estimator}) and derive from them the \emph{break length} $L^\star$, at which the quadratic term equals the linear term. Combining the cost model with the provider's pricing then gives the loss length, beyond which serving the request loses money.

\subsection{Snowball Cost}
\label{sec:decode_cost_model}
Prior work has established that autoregressive decode is memory-bandwidth-bound and that each decode step reads more KV bytes than the previous one~\cite{shazeer2019mqa, pope2023efficient, ainslie2023gqa}. We sum this per-step behavior into a request-level closed form and derive from it the threshold lengths of Section~\ref{sec:cost_model}. Serving-time latency models that guide scheduling and placement~\cite{zhong2024distserve, vidur} are fitted in the normal regime of hundreds to a few thousand output tokens, whereas we target longer outputs, where the quadratic term is a large share of decode time.

\noindent\textbf{Closed form and coefficients.}
With input length $L_{\mathrm{in}}$ and output length $L_{\mathrm{out}}$, the decode time of a single request running alone is
\begin{equation}
T_{\mathrm{decode}} = c\,L_{\mathrm{out}} + b\,(L_{\mathrm{in}} L_{\mathrm{out}}) + d\,L_{\mathrm{out}}^2 .
\label{eq:decode_closed_form}
\end{equation}
Each step re-reads the weights ($\beta_w P$ bytes, where $P$ is the number of parameters and $\beta_w$ the bytes per parameter; $\beta_w{=}2$ in BF16) and the accumulated KV cache; summing these bytes over the output gives the linear, bilinear, and quadratic terms, and dividing by the aggregate bandwidth $\mathrm{BW}\cdot G$ ($\mathrm{BW}$ is the per-GPU memory bandwidth and $G$ the number of GPUs under tensor parallelism, TP) gives
\begin{equation}
c_w = \tfrac{\beta_w P}{\mathrm{BW}\,G\,\eta_w}, \;\;
b = \tfrac{k_{\mathrm{kv}}}{\mathrm{BW}\,G\,\eta_{\mathrm{kv}}}, \;\;
d = \tfrac{k_{\mathrm{kv}}}{2\,\mathrm{BW}\,G\,\eta_{\mathrm{kv}}} ,
\label{eq:byte_coeffs}
\end{equation}
where $c$, $b$ and $d$ are the coefficients of the linear, bilinear and quadratic terms, $k_{\mathrm{kv}}$ is the per-token KV-cache footprint fixed by the model config and $\eta_w, \eta_{\mathrm{kv}} \in (0,1]$ are the achieved bandwidth utilizations of the weight and KV-cache reads (derivation, scope, and the MoE/quantization variants in Appendix~\ref{sec:cost_model_derivation}). The fitted slope also carries a per-step overhead, $c = c_w + c_{\mathrm{ovh}}$, where $c_w$ is the weight-read part; the cross-term $b$ shifts decode time by under $0.5\%$ for the short prompts of real chat data, so we drop it hereafter (Appendix~\ref{sec:cost_model_derivation}).

\noindent\textbf{Device-count scaling.}
Only the weight-read part of $c$ scales as $1/G$ while per-step overhead does not, so extra devices give a sub-linear per-request speedup (Appendix~\ref{sec:appendix_transfer_detail}). A separate memory budget (independent of the cost model) bounds the feasible output length and concurrency (Appendix~\ref{sec:cost_model_derivation}).

\subsection{Break Length and Loss Length}
\label{sec:cost_model}

\noindent\textbf{$\rho$-break length.}
We define the \emph{$\rho$-break length} as the output length at which the quadratic term reaches a fraction $\rho$ of the linear term ($d\,L^2 = \rho\,c\,L$, where we write $L$ for $L_{\mathrm{out}}$):
\begin{equation}
L_\rho = \rho\,c/d .
\label{eq:crossover}
\end{equation}
$L^\star = c/d$ is the \emph{break length}, the case $\rho = 1$, at which the quadratic term equals the linear term. Then $L_\rho = \rho\,L^\star$, and beyond $L^\star$ the quadratic term exceeds the linear term. Equivalently, a short-output linear estimate $\hat{T}=cL$ underpredicts the decode time $T$, and the relative error $(T-\hat{T})/\hat{T}=L/L^\star$ reaches $\rho$ at $L_\rho$.
Substituting Eq.~\ref{eq:byte_coeffs} with $c = c_w + c_{\mathrm{ovh}}$ splits the break length into two parts, $L^\star = (2\beta_w P/k_{\mathrm{kv}})(\eta_{\mathrm{kv}}/\eta_w) + 2\,\mathrm{BW}\,G\,\eta_{\mathrm{kv}}\,c_{\mathrm{ovh}}/k_{\mathrm{kv}}$. The first part is set by the \emph{model} (parameter bytes over per-token KV bytes) and by the \emph{engine} through its two bandwidth utilizations. The second part is the per-step overhead in units of $d$, so at fixed $\eta_{\mathrm{kv}}$ and $c_{\mathrm{ovh}}$ it grows with $G\,\mathrm{BW}$, because increasing the device count or memory bandwidth reduces $d$ while leaving $c_{\mathrm{ovh}}$ unchanged.
Section~\ref{sec:estimator} tests our cost model and locates $L^\star$ empirically.

\noindent\textbf{Where the linear regime ends.}
The break length marks a rough boundary. Below it, a linear estimate stays close to the measured decode time. Beyond it, the linear estimate falls increasingly short. Comparing $L^\star$ with the context window $L_{\max}$ shows whether a deployment reaches this regime. When $L_{\max}$ is well below $L^\star$, no output is long enough for the linear estimate to fail, and when $L_{\max}$ reaches $L^\star$, the longest outputs lie beyond the boundary (Table~\ref{tab:estimator}).

\noindent\textbf{Loss length.}
Under token-based pricing~\cite{openai2025pricing, anthropic2025pricing}, revenue scales with generated tokens while cost scales with the time the node is held. $\kappa_{\mathrm{node}}$ is the per-second cost of the serving node, $r_{\mathrm{tok}}$ is the price per output token, and $\bar{q}$ is the node's output throughput ($TPS$) over an interval of length $T$. The profit of the interval is $\pi = T(r_{\mathrm{tok}}\bar{q} - \kappa_{\mathrm{node}})$ (Appendix~\ref{sec:appendix_pricing}), so the interval is profitable only when $\bar{q}$ reaches the break-even throughput $q_{\mathrm{BE}}$:
\begin{equation}
    \bar{q} \ge q_{\mathrm{BE}} = \kappa_{\mathrm{node}}/r_{\mathrm{tok}} .
    \label{eq:profit_tps}
\end{equation}
Commercial API pricing is at most piecewise-linear, and some models are tiered by \emph{input} length~\cite{openai2025pricing, anthropic2025pricing}. As long as the per-token output price is bounded, the analysis is unchanged, because the super-linear cost is driven by \emph{output} length. On a dedicated stream, a request of length $L$ earns $r_{\mathrm{tok}}L$ and costs $\kappa_{\mathrm{node}}(cL+dL^2)$ to serve, so it is profitable only below the \emph{loss length}
\begin{equation}
L_{\mathrm{loss}} = \frac{r_{\mathrm{tok}}/\kappa_{\mathrm{node}} - c}{d} .
\label{eq:loss_len}
\end{equation}
Batching shares the weight read $c$ among concurrent requests, but each stream still reads its own KV cache, so even an unbounded number of concurrent streams only relaxes this ceiling to $L_{\mathrm{loss}}^{\infty}\!\propto\! r_{\mathrm{tok}}/(\kappa_{\mathrm{node}}\,d)$.

Section~\ref{sec:estimator} measures both parts. $d$ scales as $1/G$; the model-and-engine part differs by $13\%$ between one and four A800s while the overhead part grows with $G$; across GPU types at TP$=1$, where the overhead part is small, $L^\star$ shifts by $6\%$. The loss length depends on $G$ as well. With $\kappa_{\mathrm{node}}\propto G$ and $d\propto 1/G$ the first term of Eq.~\ref{eq:loss_len} is unchanged while the subtracted $c/d$ grows, so adding devices lowers $L_{\mathrm{loss}}$. Section~\ref{sec:estimator} calibrates $c,d$ and tabulates $L^\star$; $L_{\mathrm{loss}}$ remains in closed form, since $\kappa_{\mathrm{node}}$ and $r_{\mathrm{tok}}$ are provider-specific.

\section{Experiments}
\label{sec:experiments}
\subsection{Experimental Setup}
\noindent\textbf{Workload setup.}
\looseness=-1 To keep the workload realistic, we draw normal-request prompts from WildChat-1M~\cite{wildchat} and response lengths from the real chat distribution (Appendix~\ref{sec:appendix_workload_length} characterizes nine public datasets, and Appendix~\ref{sec:appendix_lmsys_cell} repeats a shared-serving cell with LMSYS-Chat-1M~\cite{lmsys_chat_1m} prompts). Elephant requests are mixed in at an elephant fraction $p_{\mathrm{ele}}$, their lengths set as fractions $\eta$ of the context window (Section~\ref{sec:generator}), extending beyond the length distribution of public conversation datasets into the ultra-long regime (up to 1M tokens). The prompt of an elephant request need not be long itself, which distinguishes the \textbf{long-output} problem from long-context reading~\cite{ruler}; Appendix~\ref{sec:appendix_examples} lists deployed settings where such outputs already occur.

\noindent\textbf{Length range and generation control.}
The extreme lengths we study are 128K to 1M output tokens, of the same order as today's output budgets. GPT-6.2 Astra and Claude Opus 5 allow 128K output tokens per request, and the Claude Batches API allows 300K~\cite{openai_models_docs, anthropic_models_docs}. For every elephant request, we suppress the end-of-sequence token (\texttt{ignore\_eos}), or the end-of-thinking token \texttt{</think>} for reasoning models, and set \texttt{min\_tokens} equal to \texttt{max\_tokens}, so the engine decodes exactly to the target length.

\noindent\textbf{Inference engine.} %
We use vLLM and SGLang as the production stacks under study, and HuggingFace Transformers as a controlled reference, as stated in Section~\ref{sec:inference_engine}.

\noindent\textbf{Metric collection.}
We drive each engine through its OpenAI-compatible streaming API and join each request's client-side timestamps to the server-side engine log by request ID (Appendix~\ref{sec:appendix_protocol}).

\noindent\textbf{Models.}
We evaluate chat and reasoning models spanning 7B--397B parameters and 128K--1M context windows (Table~\ref{tab:model-max-context} in Appendix~\ref{sec:appendix_models}). The official ``max output'' is a trained recommendation, not an architectural cap, so we take the context window as the effective output ceiling $L_{\max}$.

\begin{figure*}[!t]
\centering
\begin{subfigure}{0.49\textwidth}
\centering
\includegraphics[width=\linewidth]{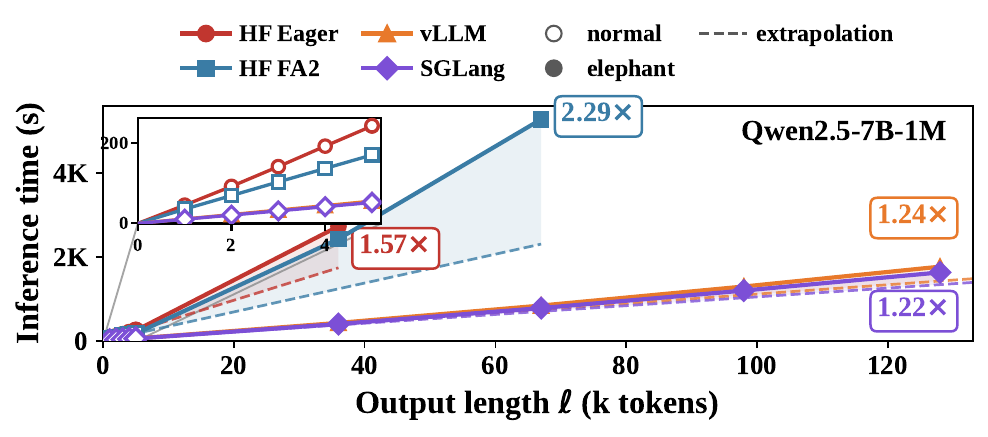}
\caption{Four engines, 1K to 128K tokens.}
\label{fig:single-user-extrapolation}
\end{subfigure}
\hfill
\begin{subfigure}{0.49\textwidth}
\centering
\includegraphics[width=\linewidth]{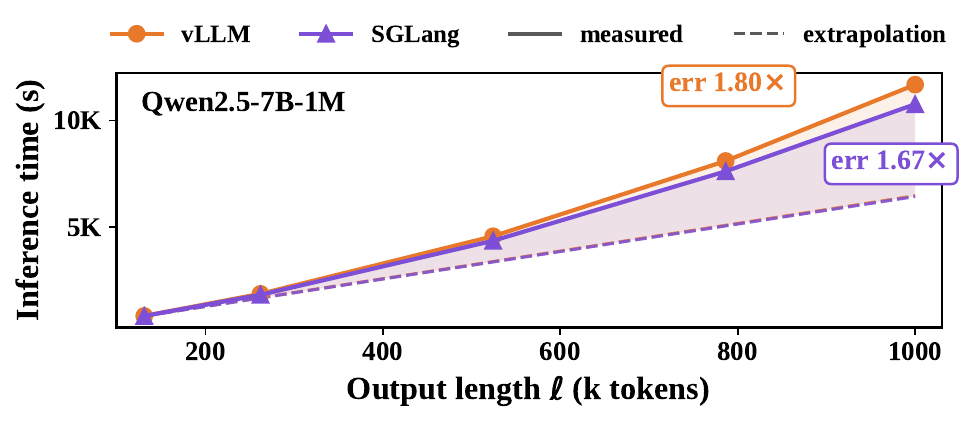}
\caption{vLLM and SGLang, 128K to 1M tokens.}
\label{fig:single-user-ultralong}
\end{subfigure}
\caption{Single-request inference time versus output length on Qwen2.5-7B-Instruct-1M. (a) vLLM, SGLang and two HuggingFace attention modes on one A100, outputs of 1K to 128K tokens; (b) vLLM and SGLang on four A800s, outputs of 128K to 1M tokens. Hollow and filled markers are normal and elephant requests. Dashed lines are linear estimates, fitted to outputs below 5K in (a) and drawn through the origin and the 131K point in (b). The boxed number is the measured time divided by the linear estimate. On every engine, inference time rises above the linear estimate as the output grows.}
\label{fig:single-user}\vspace{-0.8em}
\end{figure*}

\subsection{Elephants in Isolation}
\label{sec:single_request}
We first isolate how single-request latency scales with output length with no other traffic on the node.
We sample 5 normal requests whose prompt and response lengths come from real chat traffic ($P_{99}$ responses below ${\sim}5$K tokens; Appendix~\ref{sec:appendix_workload_length}), and elephant requests covering 5K up to 1M.

\noindent\textbf{Performance on chat models.} We first measure how inference time scales with output length for chat-model requests. On both production engines, inference time grows almost linearly through the normal range and rises above the linear trend before the context limit (Figure~\ref{fig:single-user-extrapolation}). At 128K the measured time is $1.24\times$ a short-output linear fit on vLLM and $1.22\times$ on SGLang, and at 1M the ratio grows to $1.80\times$ and $1.67\times$ (Figure~\ref{fig:single-user-ultralong}). In Figure~\ref{fig:single-user-extrapolation} the dashed lines are fitted on outputs below 5K, the forecast a provider would make from normal traffic, and in Figure~\ref{fig:single-user-ultralong} they extend the 131K point. Table~\ref{tab:baselines} calibrates its linear baseline at 131K and 262K instead, and that steeper slope reduces the 1M ratio on vLLM from $1.80\times$ to $1.66\times$. On the HuggingFace references, the ratio is already $1.57\times$ at 36K with eager attention and $2.29\times$ at 67K with FA2. FA2 has a smaller linear term, so the quadratic term is a larger fraction of its time.

\noindent\textbf{Performance on reasoning models.}
We then repeat the measurement on reasoning models. Under vLLM the scaling mirrors the chat results. The measured inference time is $1.31\times$ the linear extrapolation at 256K tokens (Qwen3-235B-A22B-Thinking, TP$=$8) and $1.37\times$ at 118K (gpt-oss-120b, TP$=$4) (Figure~\ref{fig:single-user-reasoning}, Appendix~\ref{sec:appendix_reasoning_fig}). 

\begin{tcolorbox}[colback=gray!10, colframe=gray!50!black, title=\textbf{Insight 1}]
\looseness=-1 A cost estimate fitted on normal-length outputs underestimates the decode time of ultra-long outputs, even on optimized production engines such as vLLM and SGLang. The gap grows with the output length, and the length at which it becomes large differs across LLMs, serving engines and hardware. 
\end{tcolorbox}

\subsection{Calibrating and Transferring the Cost Model}
\label{sec:estimator}
We now fit the cost model of Section~\ref{sec:modeling} to these measurements and report its coefficients and break lengths. We then test whether coefficients calibrated on one deployment predict other deployments.

\noindent\textbf{Coefficient fit.}
Fitting Eq.~\ref{eq:decode_closed_form} to the Figure~\ref{fig:single-user} deployment with repetition penalty $1.0$ (Qwen2.5-7B-1M, vLLM, $G{=}4$; $11$ observations at $9$ lengths from $512$ to $262$K, $R^2=0.999999$) gives $c=5.70{\times}10^{-3}$~s/tok and $d=5.65{\times}10^{-9}$~s/tok$^2$: $L^\star=1.01$M lies at the edge of the context window. On eight A800s, Qwen2.5-72B has a fitted break length of $L^\star=1.17$M, of the same order as the architectural estimate $2\beta_w P/k_{\mathrm{kv}}\approx8.9{\times}10^{5}$. Table~\ref{tab:estimator} spans models, GPUs, and engines, with $L_{\max}/L^\star$ at $0.11$ for 72B, below the break length, and $1.0$ for 7B-1M, at the break length.

\noindent\textbf{Baselines with matched calibration.}
Table~\ref{tab:baselines} compares our model with a linear model, a general quadratic and a Vidur-style random forest on the vLLM 7B-1M sweep. Calibrated at $131$K and $262$K, our model is within $4.3\%$ at $1$M, whereas the linear model underestimates decode time there by $39.7\%$. A general quadratic $a_0+a_1L+a_2L^2$ needs a third calibration point; at $524$K, it comes within $0.6\%$ at $1$M, but its intercept $a_0$ fits to $23.3$, $-0.08$ and $49.5$\,s on the three deployments. Given the same three points, our model is within $1.4\%$ at $1$M and $0.2\%$ at $65$K. A Vidur-style random forest~\cite{vidur} underestimates decode time at $1$M by $51.7\%$, or by $35.2\%$ with our two calibration points added, because it cannot predict beyond its longest training length.

\noindent\textbf{The break length.}
\looseness=-1 The KV term's share of decode time, $(L/L^\star)/(1+L/L^\star)$, is about $0.5\%$ at the $5$K outputs covering most real traffic, $12\%$ at $131$K, $34\%$ at $524$K, and $50\%$ at $1$M on vLLM with the coefficients of Table~\ref{tab:estimator} ($47\%$ at $1$M on SGLang). On the same 72B model and four A800s, $L^\star$ shifts from $29$K to $1.26$M when the engine switches from HuggingFace eager attention to SGLang (Table~\ref{tab:estimator}). From one to four A800s, $L^\star$ of 7B-1M on vLLM grows from $560$K to $1.01$M. On one card, $L^\star$ shifts by $6\%$ from the A800 to the H800. Of the five 72B configurations, only HuggingFace eager attention has $L_{\max}/L^\star>1$, so a short-output estimate already underestimates its decode time within normal output lengths.

\noindent\textbf{Prediction in unseen settings.}
In Equation~\ref{eq:byte_coeffs}, $P$ and $k_{\mathrm{kv}}$ come from the model config and $\mathrm{BW}$ from the GPU datasheet, so the engine is described by $\eta_w$, $\eta_{\mathrm{kv}}$ and $c_{\mathrm{ovh}}$. Table~\ref{tab:heldout} calibrates the cost model on one deployment and predicts a setting that differs in one factor. Calibrated at $131$K and $262$K, the cost model predicts decode time at $1$M within $4.3\%$ on vLLM and $0.2\%$ on SGLang. Rescaled by the datasheet bandwidth, the coefficients of one A800 predict the decode time of one A100 within $1.5\%$ at $262$K, and those of the H800 within $4.7\%$ from $8$K to $131$K. Only $d$ follows $1/G$, so across GPU counts we predict $d$, which comes within $11.3\%$ from one to four A800s and $9.6\%$ from four to eight. Across model sizes, $\eta_{\mathrm{kv}}$ ranges from $0.63$ to $0.72$ on five vLLM deployments of 7B-1M and 72B, so calibrating any one predicts $d$ on the others within $13\%$ (Appendix~\ref{sec:appendix_transfer_detail}). On hybrid attention, with $k_{\mathrm{kv}}$ counting only the layers whose cache grows with length, the cost model predicts gpt-oss-120b within $1.9\%$ at $118$K and Qwen3.5-397B-A17B within $0.9\%$ at $131$K (Appendix~\ref{sec:appendix_hybrid}).

\begin{table}[t]
\centering
\caption{Calibrated coefficients and threshold lengths. vLLM rows are measured with the default repetition penalty of Qwen2.5 disabled; HuggingFace rows keep the default sampler. $c$ is in ms per token, $d$ in ns per token$^2$, and $L_{\max}$ is the context window.}
\label{tab:estimator}
\small
\setlength{\tabcolsep}{2.8pt}
\begin{tabular}{@{}llrrrr@{}}
\toprule
\textbf{Model} & \textbf{GPU, engine} & $c$ & $d$ & $L^\star$ & $\frac{L_{\max}}{L^\star}$ \\
\midrule
\multirow{5}{*}{72B}
 & A800$\times$4 HF Eager & 114 & 3{,}900 & 29K & 4.5 \\
 & A800$\times$4 HF FA2   & 119 & 481 & 247K & 0.53 \\
 & A800$\times$4 SGLang   & 33.0 & 26.2 & 1.26M & 0.10 \\
 & A800$\times$4 vLLM     & 31.4 & 29.1 & 1.08M & 0.12 \\
 & A800$\times$8 vLLM     & 18.8 & 16.1 & 1.17M & 0.11 \\
\midrule
\multirow{4}{*}{7B-1M}
 & A100$\times$1 vLLM   & 11.0 & 21.8 & 507K & 1.99 \\
 & A800$\times$1 vLLM   & 11.2 & 20.1 & 560K & 1.80 \\
 & A800$\times$4 vLLM   & 5.70 & 5.65 & 1.01M & 1.0 \\
 & A800$\times$4 SGLang & 5.67 & 5.09 & 1.11M & 0.91 \\
\bottomrule
\end{tabular}
\vspace{-0.8em}
\end{table}

\begin{table}[t]
\centering
\caption{Prediction error (\%, predicted vs.\ measured) on vLLM 7B-1M ($4\times$A800) for four predictors, each given the calibration points listed (cal.). Vidur RF is the Vidur-style random forest, Quadratic is $a_0{+}a_1L{+}a_2L^2$, and ours is $cL{+}dL^2$.}
\label{tab:baselines}
\small
\setlength{\tabcolsep}{2.6pt}
\begin{tabular}{@{}llrrrr@{}}
\toprule
\textbf{Predictor} & \textbf{Calibration (K)} & \textbf{65K} & \textbf{524K} & \textbf{786K} & \textbf{1M} \\
\midrule
Linear $cL$ & 131, 262 & $+16.1$ & $-19.4$ & $-31.6$ & $-39.7$ \\
Vidur RF & $\le$4 & $-6.9$ & $-35.4$ & $-45.2$ & $-51.7$ \\
Vidur RF & $\le$4, 131, 262 & $-6.9$ & $-16.9$ & $-27.5$ & $-35.2$ \\
Quadratic & 131, 262, 524 & $+3.1$ & cal. & $+0.1$ & $-0.6$ \\
Ours & 131, 262 & $+0.9$ & $-1.5$ & $-2.8$ & $-4.3$ \\
Ours & 131, 262, 524 & $-0.2$ & cal. & $-0.4$ & $-1.4$ \\
\bottomrule
\end{tabular}
\vspace{-0.8em}
\end{table}

\begin{table}[t]
\centering
\caption{Prediction in unseen settings. Each row calibrates the cost model on one deployment (vLLM unless noted; outputs of 131K and 262K, or two shorter lengths for the attention rows) and predicts the decode time or $d$ in a setting that differs in one factor.}
\label{tab:heldout}
\small
\setlength{\tabcolsep}{3pt}
\begin{tabular}{@{}lllr@{}}
\toprule
\textbf{Unseen} & \textbf{Calibrated on} & \textbf{Predicted} & \textbf{Error} \\
\midrule
\multirow{3}{*}{Length} & 7B-1M, A800$\times$4 & time at 1M & $-4.3\%$ \\
 & same, SGLang & time at 1M & $-0.2\%$ \\
 & 7B-1M, A100$\times$1 & time at 786K & $-2.8\%$ \\
\midrule
\multirow{2}{*}{GPU type} & 7B-1M, A800$\times$1 & A100, 262K & $-1.5\%$ \\
 & 7B-1M, A800$\times$1 & H800, 131K & $+4.7\%$ \\
\midrule
\multirow{2}{*}{GPU count} & 7B-1M, A800$\times$1 & $d$, A800$\times$4 & $-11.3\%$ \\
 & 72B, A800$\times$4 & $d$, A800$\times$8 & $-9.6\%$ \\
\midrule
Model & 7B-1M, A800$\times$4 & $d$ of 72B & $+11.0\%$ \\
\midrule
\multirow{2}{*}{Attention} & gpt-oss-120b & time at 118K & $-1.9\%$ \\
 & Qwen3.5-397B & time at 131K & $-0.9\%$ \\
\bottomrule
\end{tabular}
\vspace{-0.8em}
\end{table}

Appendix~\ref{sec:appendix_transfer_detail} compares the KV-read efficiency of the two engines.

\noindent\textbf{Batched form.} Appendix~\ref{sec:appendix_batch_tput} extends the cost model to batches of concurrent requests. Fitted on six concurrency levels at three output lengths, it reproduces all $18$ cells within $2.8\%$ and predicts unseen $65$K cells within $1.0\%$.

\subsection{Impact of Elephant Requests in Shared Serving}
\label{sec:mixed}
\begin{figure*}[t]
\centering
\includegraphics[width=0.98\textwidth]{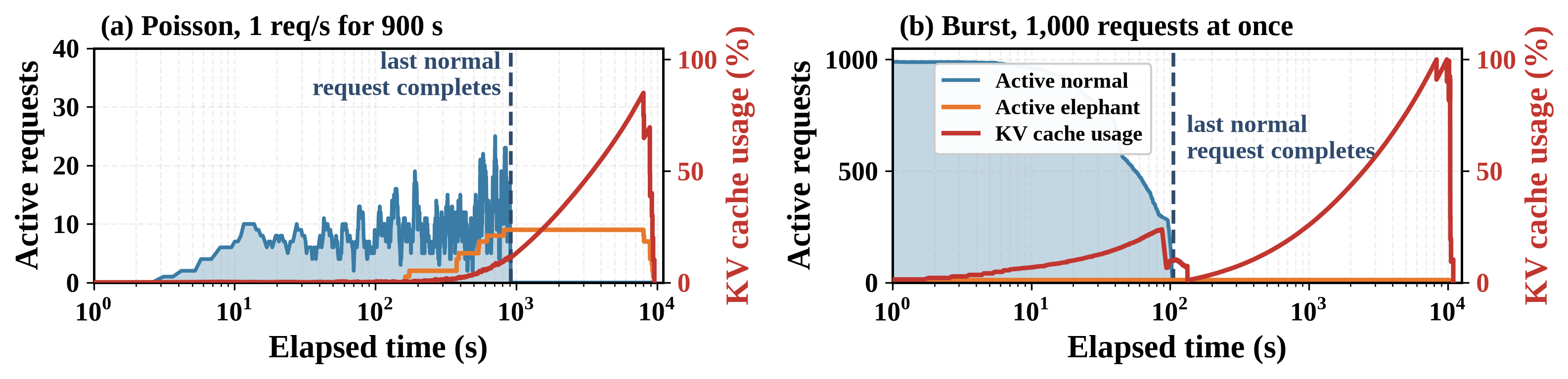}
\caption{Active normal and elephant requests (left axis) and KV-cache usage (right axis) with $1\%$ elephant requests on vLLM (Qwen2.5-72B-Instruct, TP$=8$), under (a) Poisson arrival at $1$\,req/s for $900$\,s, which yields $901$ requests with $9$ elephants, and (b) a burst of $1{,}000$ requests with $11$ elephants. In both cases KV-cache usage peaks after the last normal request completes, when only the elephants are still decoding.}
\label{fig:kv}\vspace{-1.0em}
\end{figure*}

\label{sec:multi_user}

\looseness=-1 Production serving is shared, so we now measure how elephant requests affect normal traffic and the server's resource usage. We send mixed workloads of normal and elephant requests to vLLM, which serves Qwen2.5-72B-Instruct on an A800 node. Unless otherwise stated, each cell contains 1{,}000 requests, normal requests draw prompt lengths from WildChat, and elephant requests use a controlled output length.

\begin{table}[t]
\centering
\caption{Per-class median latency under two arrival processes (Qwen2.5-72B, vLLM, TP=8, WildChat). Poisson draws 1 req/s over a 900 s window, which yields 901 requests, 9 of them elephants; burst launches $1{,}000$ requests at once, 11 of them elephants. The percentage after each class is the elephant fraction; the last two columns give the share that meets each latency bound.}
\label{tab:slo_single}
\small
\setlength{\tabcolsep}{3.5pt}
\begin{tabular}{@{}llrrrr@{}}
\toprule
 & & & & \multicolumn{2}{c}{\textbf{SLO met}} \\
\cmidrule(l){5-6}
\textbf{Arrival} & \textbf{Class} & \textbf{TTFT} & \textbf{E2E} & \textbf{$\le$2\,s} & \textbf{$\le$60\,s} \\
\midrule
\multirow{3}{*}{Poisson}
 & Normal, 0\%    & 0.10\,s & 11.3\,s & 100\% & 100\% \\
 & Normal, 1\%   & 0.12\,s & 11.7\,s & 100\% & 100\% \\
 & Elephant, 1\% & 0.12\,s & 8{,}541\,s & 100\% & 0\% \\
\midrule
\multirow{3}{*}{Burst}
 & Normal, 0\%    & 10.4\,s & 54.2\,s & 30.0\% & 53.2\% \\
 & Normal, 1\%   & 12.2\,s & 57.5\,s & 29.9\% & 51.9\% \\
 & Elephant, 1\% & 8.1\,s & 10{,}439\,s & 27.3\% & 0\% \\
\bottomrule
\end{tabular}
\vspace{-0.8em}
\end{table}

\subsubsection{Impact on Users}
\label{sec:ablation_main}
\begin{figure}[t]
\centering
\includegraphics[width=0.99\columnwidth]{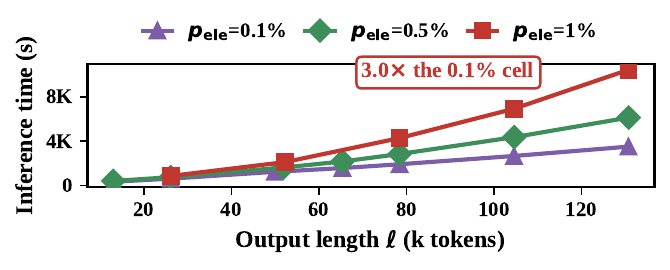}
\caption{Inference time of elephant requests versus their output length when they share a node with normal requests. Each curve is one elephant fraction among 1{,}000 requests arriving in a burst (Qwen2.5-72B-Instruct, vLLM, TP$=8$).}
\label{fig:mixed_vs_single}
\vspace{-1.0em}
\end{figure}

\noindent\textbf{Impact on elephant requests.}
In shared serving an elephant's inference time retains the quadratic scaling it has in isolation (Figure~\ref{fig:mixed_vs_single}). The inference time also grows with the elephant fraction, and every curve rises above the linear trend as the output grows. Relative to the $0.1\%$ cell, an elephant in the $1\%$ cell takes $2.24\times$ as long at 78K.
At 128K the slowdown grows with the number of elephants on the node. In the $0.5\%$ cell, six elephants fill $62\%$ of the KV cache, and each takes $1.75\times$ its time in the $0.1\%$ cell. In the $1\%$ cell, eleven elephants exceed the KV cache, so they cannot all decode at once, and the ratio reaches $2.98\times$.

\noindent\textbf{Effect of elephants on normal requests.}
Under Poisson arrival, requests do not queue, and the elephants slow normal requests slightly (Table~\ref{tab:slo_single}). Under the burst workload, elephants do not measurably slow normal requests. Across elephant fractions from $0$ to $5\%$ ($0$ to $46$ elephants), the median normal TTFT stays between $8.6$ and $12.5$\,s with no trend in the elephant fraction, and normal per-token decode is flat within $2.5\%$, while elephant latencies range over orders of magnitude (Figure~\ref{fig:ablation_plong_main}). Normal requests finish during the early high-concurrency phase, before the elephants' KV cache grows.

\begin{tcolorbox}[colback=gray!10, colframe=gray!50!black, title=\textbf{Insight 2}]
Shared serving affects the elephant requests and the provider far more than the normal requests. Each elephant request generates every token more slowly as its output grows, and more slowly still when more elephant requests share the server. The provider's GPUs stay occupied by a few elephant requests long after the normal traffic has finished. Normal requests keep a fast first token and their SLO attainment barely changes, while those that arrive after the elephants' responses have grown take slightly longer to finish.
\end{tcolorbox}

\subsubsection{Impact on the Server System: SLO, Concurrency, and Memory}
\label{sec:system_slo_concurrency}

\noindent\textbf{SLO.} We evaluate every request against $\mathrm{TTFT}\!\le\!2$\,s and $\mathrm{E2E}\!\le\!60$\,s (Table~\ref{tab:slo_single}). Under Poisson arrival, every normal request satisfies both SLOs, with or without elephants. For the same normal requests, the median completion time is $11.2$\,s with elephants versus $11.3$\,s without them when they arrive in the first three minutes, and $17.6$\,s versus $11.3$\,s in the last three. The median TTFT of these late requests rises only from $0.10$ to $0.17$\,s. Their decode steps also read the accumulated KV caches of nine elephants, so each step takes longer as those caches grow. Under burst arrival, most normal requests miss these bounds because they queue behind the other requests, with or without elephants. Adding $1\%$ elephants changes the normal TTFT pass rate from $30.0\%$ to $29.9\%$ and the E2E pass rate from $53.2\%$ to $51.9\%$. With $1\%$ elephants, the SLO attainment of normal requests barely changes under either arrival process, whereas normal requests that arrive while the elephants are decoding take longer to finish.

\noindent\textbf{Concurrency and memory.} In the $1\%$ cell, normal traffic clears in ${\sim}105$\,s while the 11 elephants take ${\sim}10.9$K\,s to finish decoding, and KV usage climbs toward $100\%$ (Figure~\ref{fig:kv}).

\subsubsection{Impact on Providers}
\noindent\textbf{Cost after the normal requests complete.}
Under Poisson arrival with $1\%$ elephants, the node produces about $440$ output tokens per second until the last normal request completes at $913$\,s, and about $130$ afterwards, while $9$ elephants decode for another $8{,}700$\,s. In the $1\%$-elephant burst, the node produces about $3{,}700$ output tokens per second while the normal requests run. After they complete at $105$\,s, the $11$ remaining elephants produce about $130$ tokens per second, because each decode step emits only one token per running request. In both cases the node operates at a loss over the elephant-only interval whenever $q_{\mathrm{BE}}$ exceeds that rate.

\noindent\textbf{The loss length under shared serving.}
Section~\ref{sec:cost_model} gives the output length past which a single request becomes unprofitable. Under shared serving, profit depends on the output throughput of the whole node. Every decode step reads the model weights once for the whole batch but each request's cache separately, so on vLLM $64$ concurrent requests produce $42\times$ the throughput of a single request at $2$K outputs and only $22.5\times$ at $32$K (Appendix~\ref{sec:appendix_batch_tput}). Late in a burst only elephants remain, so the node's throughput drops to what they produce, and the node falls below the break-even rate at a shorter output length than it would with a full batch.

\begin{tcolorbox}[colback=gray!10, colframe=gray!50!black, title=\textbf{Insight 3}]
A provider's revenue grows piecewise-linearly with the number of output tokens, whereas its cost grows with GPU time, which grows super-linearly with the response length. Beyond a threshold response length, a request therefore costs more to serve than it earns. 
\end{tcolorbox}

\section{Conclusion}
\looseness=-1 We presented \textsc{SnowballCost}, a framework that measures how ultra-long responses affect inference. A linear cost estimate fitted at normal output lengths underestimates the decode time of long responses. 
When normal and elephant requests share a server, the elephant requests generate tokens more slowly as their output grows and keep the GPUs occupied long after the normal traffic ends, while normal requests are only slightly delayed and their SLO attainment barely changes. Beyond a length given by our cost model, a response becomes unprofitable.

\section{Limitations}
Our evaluations run on single-node multi-GPU servers. Cluster-level effects such as inter-node communication, cross-node pipeline parallelism and global load balancing are outside these measurements.

We do not design or evaluate scheduling or admission-control policies, and we leave them to future work.

\section{Ethics Statement}
This work adheres to the ACL Ethics Policy. We study the impact of ultra-long LLM responses in LLM inference services. Our experiments use controlled resource-intensive workloads to measure serving behavior under long-output traffic and to inform more reliable serving systems. All datasets utilized in our experiments are publicly available benchmarks and contain no human subjects or personally identifiable information. To support reproduction and reuse, we release the workload specification, the request traces and per-request logs behind every reported cell, the log schema, and the analysis and fitting scripts at \url{https://github.com/fengxijia/SnowballCost}.

\bibliography{references}

\newpage
\appendix
\input{appendix}

\end{document}

%% file: appendix.tex
\section{Derivation of the Cost Model}
\label{sec:cost_model_derivation}

\noindent\textbf{Summing the per-step bytes.}
Each decode step $t$ is memory-bandwidth-bound: the engine reads the full weights $W=\beta_w P$ bytes ($\beta_w$ bytes per parameter; $2$ in BF16) plus a KV cache of $(L_{\mathrm{in}}+t)\,k_{\mathrm{kv}}$ bytes, where $k_{\mathrm{kv}}$ is the per-token KV footprint, taken from the model config as $k_{\mathrm{kv}} = 2\,\beta_{\mathrm{kv}}\,n_{\mathrm{layers}}\,n_{\mathrm{kv\text{-}heads}}\,d_{\mathrm{head}}$ with $\beta_{\mathrm{kv}}$ bytes per element and, under GQA~\cite{ainslie2023gqa}, the \emph{key/value} head count rather than the query head count. Summing the bytes moved over $t=0,\dots,L_{\mathrm{out}}-1$ and using $\sum_{t=0}^{L_{\mathrm{out}}-1}t=L_{\mathrm{out}}(L_{\mathrm{out}}-1)/2$ gives a linear weight term ($\beta_w P\,L_{\mathrm{out}}$), a bilinear term ($k_{\mathrm{kv}}\,L_{\mathrm{in}}L_{\mathrm{out}}$), and a quadratic KV term ($k_{\mathrm{kv}}\,L_{\mathrm{out}}^2/2$), plus a leftover linear piece $-k_{\mathrm{kv}}L_{\mathrm{out}}/2$; the latter amounts to $k_{\mathrm{kv}}/2$ bytes per token, about $10^{-6}$ of the weight read, and is absorbed into the fitted $c$. Dividing by the aggregate bandwidth $\mathrm{BW}\cdot G$ (and folding in the achieved utilizations $\eta_w,\eta_{\mathrm{kv}}$) yields the coefficients of Eq.~\ref{eq:byte_coeffs}, and equating the linear and quadratic contributions, $cL=dL^2$, gives the break length $L^\star=c/d$ of Eq.~\ref{eq:crossover}.

\noindent\textbf{Scope.} The weight term assumes a dense model: mixture-of-experts replaces $P$ by the activated parameters plus routing cost, while quantized weights or KV enter through $\beta_w$ and $\beta_{\mathrm{kv}}$. The input cross-term $b$ is negligible in our setting. Across the input lengths we test, which follow the real chat distribution and stay well below the output lengths of interest, it shifts decode time by under $0.5\%$, so the main text omits it. Workloads that pair very long inputs with long outputs need to keep it.

\noindent\textbf{Three refinements of Eq.~\ref{eq:byte_coeffs}.} First, when a model has fewer KV heads than tensor-parallel devices, each device holds a replicated head, so the KV term scales with $\min(G, n_{\mathrm{kv\text{-}heads}})$ instead of $G$. The Table~\ref{tab:estimator} configurations are unaffected; Qwen3-235B (4 heads, TP$=8$) and Qwen3.5-397B (2 heads, TP$=8$) are affected. Second, on hybrid attention only the layers whose cache grows with length enter $k_{\mathrm{kv}}$. For gpt-oss-120b these are its $18$ full-attention layers, while its other $18$ layers use a $128$-token sliding window; for Qwen3.5-397B they are its $15$ full-attention layers, while its other $45$ layers use linear attention with a fixed-size state. Third, the two calibration points must lie where the quadratic term exceeds the per-step jitter of the deployment, about $4\%$ in our data. Calibrating at $1$K and $2$K tokens, where the term is below $1\%$, returns a negative $d$.

\noindent\textbf{KV-memory feasibility.}
Equation~\ref{eq:decode_closed_form} sets the \emph{time} of a request; a separate memory budget sets the \emph{feasible region}. $\mathrm{Cap}$ is the memory of one GPU and $u$ is the fraction usable for weights and KV cache. The longest sequence that fits in memory is
\begin{equation}
L_{\mathrm{mem}} = \frac{\mathrm{Cap}\,G\,u - \beta_w P}{k_{\mathrm{kv}}} ,
\label{eq:kv_single}
\end{equation}
while concurrent requests must jointly satisfy the KV-saturation bound $\sum_i (L_{\mathrm{in},i} + L_{\mathrm{out},i})\,k_{\mathrm{kv}} \le \mathrm{Cap}\,G\,u - \beta_w P$, which caps the achievable concurrency once outputs grow long. These constraints bound admissibility and concurrency but do not enter the time model.

\section{Revenue, Cost, and Profit under Token Pricing}
\label{sec:appendix_pricing}

\noindent\textbf{Revenue.} Under token-based pricing, revenue $R$ is a function of the processed token count. In the simplest linear billing model it is directly proportional to the number of generated tokens, and thus approximately proportional to the response length, $R \propto l$.

\noindent\textbf{Cost.} We model provider cost as proportional to serving time. Let $\kappa_{\mathrm{node}}$ denote the amortized per-second operating cost of keeping the serving node allocated (power plus hardware rental or amortization and operations), $r_{\mathrm{tok}}$ the revenue per generated token, $N_{\mathrm{out}}$ the number of generated tokens in an interval of length $T$, and $\bar{q}=N_{\mathrm{out}}/T$ the average output-token rate.

\noindent\textbf{Profit.} Combining the two gives
\begin{equation}
    \pi = r_{\mathrm{tok}}N_{\mathrm{out}} - \kappa_{\mathrm{node}}T
        = T(r_{\mathrm{tok}}\bar{q} - \kappa_{\mathrm{node}}),
\end{equation}
from which the break-even condition of Eq.~\ref{eq:profit_tps} follows. The formulation separates two cases. For an isolated request whose latency grows super-linearly with output length, the cost per generated token rises with length. For a shared serving system the relevant quantity is instead the interval-level $\bar{q}$: if other requests keep the node productively generating tokens, they offset the cost of the same wall-clock interval; if the system enters a slow elephant tail, profit turns negative whenever $\bar{q}<q_{\mathrm{BE}}$.

\section{Single-Request Scaling on Reasoning Models}
\label{sec:appendix_reasoning_fig}

\begin{figure}[h]
\centering
\includegraphics[width=0.99\columnwidth]{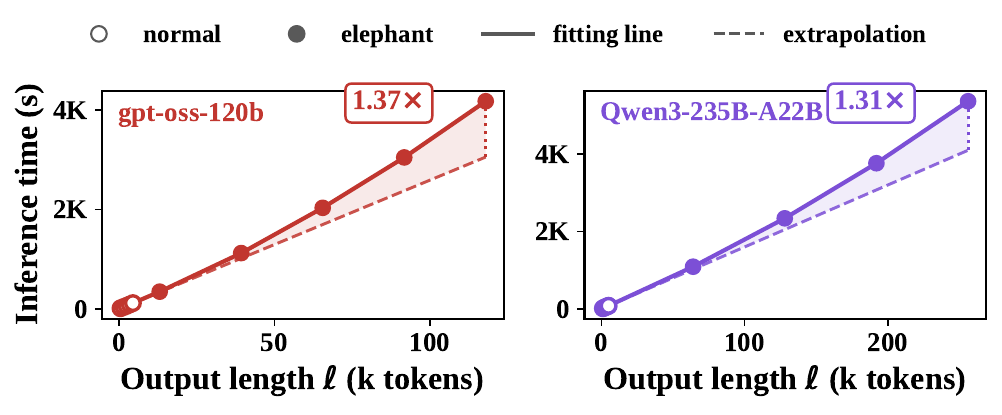}
\caption{Single-request inference time vs.\ output length for reasoning models on vLLM. Hollow markers denote the normal requests, filled markers the elephant requests. Dashed lines extrapolate the normal-range linear fit to the elephant range.}
\label{fig:single-user-reasoning}
\end{figure}

Figure~\ref{fig:single-user-reasoning} repeats the single-request sweep of Section~\ref{sec:single_request} on two reasoning models under vLLM. On both models, inference time rises above the linear trend past the normal range, as on the chat models. The measured time is $1.31\times$ the normal-range linear fit at $256$K on Qwen3-235B-A22B-Thinking and $1.37\times$ at $118$K on gpt-oss-120b.

\section{Evaluated Models}
\label{sec:appendix_models}

Table~\ref{tab:model-max-context} lists the seven models we serve, their type, and the two length limits that matter for the elephant definition. The context window column is the context limit we serve, which bounds prompt plus output.

\begin{table}[h]
  \caption{Evaluated models. Max output is the model card's recommendation, which the engines do not enforce, so $L_{\max}$ is the context window.}
  \label{tab:model-max-context}
  \centering
  \resizebox{\linewidth}{!}{
  \begin{tabular}{llrr}
    \toprule
    \textbf{Model} & \textbf{Type} & \textbf{Max context} & \textbf{Max output} \\
    \midrule
    Qwen2.5-7B-Instruct-1M   & Chat      & 1{,}010{,}000 & 8{,}192 \\
    Qwen2.5-14B-Instruct     & Chat      & 131{,}072 & 8{,}192 \\
    Qwen2.5-32B-Instruct     & Chat      & 131{,}072 & 8{,}192 \\
    Qwen2.5-72B-Instruct     & Chat      & 131{,}072 & 8{,}192 \\
    gpt-oss-120b             & Reasoning & 131{,}072 & 131{,}072 \\
    Qwen3-235B-A22B-Thinking & Reasoning & 262{,}144 & Not specified \\
    Qwen3.5-397B-A17B-FP8    & Reasoning & 131{,}072 & Not specified \\
    \bottomrule
  \end{tabular}}
\end{table}

\section{Engine and Hardware Transfer in Detail}
\label{sec:appendix_transfer_detail}

\noindent\textbf{KV-read efficiency of each engine.} Engines differ in $d$. Using Equation~\ref{eq:byte_coeffs}, we express the difference as a KV-read efficiency, $\eta_{\mathrm{kv}}=k_{\mathrm{kv}}/(2\,\mathrm{BW}\,G\,d)$. To compare the two production engines on the same sampling, we disable the repetition penalty of $1.05$ that vLLM takes from the Qwen2.5 generation config by default. With matched sampling on 7B/A800, $\eta_{\mathrm{kv}}$ is $0.63$ for vLLM and $0.70$ for SGLang, and the two values of $d$ lie within $11\%$ of each other. The attention backend also changes $\eta_{\mathrm{kv}}$. The gpt-oss-120b run falls back to Triton attention and fits $\eta_{\mathrm{kv}}=0.027$, far below the FlashAttention range, so a changed backend needs its own calibration.

\noindent\textbf{Effect of the serving stack on $d$.}
On one model and one node, the serving stack shifts $d$ across two orders of magnitude, and with it the break length (Table~\ref{tab:estimator}). Eager attention gives $d=3.9{\times}10^{-6}$ and $L^\star\!\approx\!29$K, so its context window spans $4.5$ break lengths and a short-output estimate fails inside the normal operating range; FlashAttention-2 alone cuts $d$ by $8\times$, and the production stacks cut it by two orders of magnitude. Two differences prevent a direct kernel comparison. The HuggingFace rows replicate the full KV cache across their four devices while the production rows shard it. Profiling the FlashAttention-2 configuration at 131K attributes only $5.6$\,ms per step to attention, compared with ${\sim}17$\,ms for concatenating the cache. The quadratic term therefore covers the whole per-step KV handling, not the attention kernel alone. The linear coefficient stays within $4\%$ between the two HuggingFace variants ($0.114$ versus $0.119$\,s/tok), which places their difference on the quadratic term rather than on general runtime efficiency.

\noindent\textbf{Transfer across deployments.} The KV-read utilization of vLLM 0.11.0 is $0.715$ and $0.634$ on 7B-1M with one and four A800s, $0.659$ on 7B-1M with one A100, and $0.704$ and $0.636$ on 72B with four and eight A800s ($k_{\mathrm{kv}}=327{,}680$ bytes per token for 72B). Calibrating $\eta_{\mathrm{kv}}$ on any one of the five and predicting $d$ on the other four falls within $13\%$, across a $10\times$ change in model size and an $8\times$ change in device count. 

\noindent\textbf{GPU type.} The A800 against H800 comparison of Section~\ref{sec:estimator} serves Qwen2.5-7B-1M on vLLM 0.19.1 with FlashAttention at TP$=1$ on one card of each type and sweeps $8$K to $131$K. Five-point fits give $c=1.035{\times}10^{-2}$ and $5.97{\times}10^{-3}$\,s/tok and $d=1.82{\times}10^{-8}$ and $9.89{\times}10^{-9}$\,s/tok$^2$, ratios of $0.577$ and $0.542$ against the datasheet bandwidth ratio of $0.597$; the residual is the higher KV-read utilization of the H800 ($\eta_{\mathrm{kv}}=0.87$ against $0.79$). Two-point calibration at $32$K and $64$K predicts the $131$K point within $0.3\%$ on both cards.

\section{Experimental Protocol}
\label{sec:appendix_protocol}

\noindent\textbf{Measurement harness.} Each engine is driven through its OpenAI-compatible streaming API with a pool of concurrent workers. The client timestamps submission, first-token arrival, and stream completion for every request, which gives the TTFT, TBT, and E2E latencies of Section~\ref{sec:monitor}. Each request carries a unique id that joins its client trace to the server-side engine log, so client- and server-side metrics are matched per request. Server-side concurrency and KV-cache series are polled from the engine's own metrics endpoint.

\noindent\textbf{Software.} The main vLLM measurements use vLLM 0.11.0; the GPU-type transfer, the Qwen3.5-397B sweep, and the LMSYS-Chat-1M cell use vLLM 0.19.1, as their sections state. vLLM 0.11.0 runs with the V1 engine (\texttt{VLLM\_USE\_V1=1}) under Python 3.10 and NCCL 2.27.3. vLLM selects the FlashAttention backend on V1 for the Qwen2.5 and Qwen3 models; for gpt-oss-120b it falls back to the Triton backend, which we note because it changes the attention implementation under comparison. SGLang runs use its default FlashInfer attention and sampling backends. The HuggingFace baselines set \texttt{attn\_implementation} to \texttt{eager} and \texttt{flash\_attention\_2} respectively. The vLLM environment is \texttt{vllm}~$0.11.0$ on \texttt{torch}~$2.8.0{+}\texttt{cu}129$ with \texttt{transformers}~$4.57.3$ and \texttt{flash-attn}~$2.8.3$; the SGLang environment is \texttt{sglang}~$0.5.3$ on \texttt{torch}~$2.8.0$ with \texttt{transformers}~$4.56.1$ and \texttt{flash-attn}~$2.8.0$.

\noindent\textbf{Serving configuration.} Unless stated otherwise, models are served in \texttt{bfloat16} with no quantization, \texttt{kv\_cache\_dtype=auto}, chunked prefill enabled (\texttt{max\_num\_batched\_tokens}$=8192$), prefix caching enabled, and CUDA graphs enabled (\texttt{enforce\_eager=False}). Table~\ref{tab:protocol} lists the per-experiment tensor-parallel width, GPU allocation, and context limit. The 72B mixed-workload node reports a KV cache of $1{,}266{,}720$ tokens, or $9.66$ full-length contexts. For Qwen2.5-72B, we enable YaRN rope scaling (factor $4.0$) to extend its native $32$K context to $128$K.

\noindent\textbf{Workload and sampling.} Decoding is greedy ($\text{temperature}=0$) for all single-request and mixed-workload cells; for reasoning models the visible answer segment is sampled at temperature $0.7$ with a $1024$-token budget. Single-request cells issue one request per length point sequentially at concurrency $1$, after a two-request server warmup. Mixed-workload cells issue $N=1000$ requests per cell under a burst launch with concurrency $500$ and seed $2025$; the realized elephant counts are $2$, $6$, $11$, $21$ and $46$ for $p_{\mathrm{ele}}=0.001$, $0.005$, $0.01$, $0.02$ and $0.05$. Mixed-workload cells were run once each without a separate warmup phase, so the first requests of a burst include the server's warmup time; the arrival ablation of Appendix~\ref{sec:appendix_arrival_ablation} isolates this effect. The ultra-long $128$K--$1$M sweep disables dual-chunk attention on both engines so that the two stacks run the same attention path. For chat models, we set \texttt{ignore\_eos} and \texttt{min\_tokens} equal to \texttt{max\_tokens} on every elephant request, so the engine decodes exactly to the target length; every elephant in the single-request sweeps finishes with \texttt{finish\_reason} \texttt{length}. The $128$K--$1$M sweeps and the length sweeps with repetition penalty $1.0$ send the single user turn \texttt{Write an extremely long, detailed technical essay.} The single-request length sweeps up to $131$K take the shortest prompts of the sampled chat dataset, and in the mixed-workload cells an elephant keeps the dataset prompt drawn for its slot, so elephants and normal requests share one prompt distribution. The single-request and batched measurements that compare vLLM with SGLang (Sections~\ref{sec:single_request} and~\ref{sec:estimator}, Table~\ref{tab:batch_tput}) set \texttt{repetition\_penalty} to $1.0$ on vLLM, so both engines sample identically; the HuggingFace rows in Table~\ref{tab:estimator} and the shared-serving cells of Section~\ref{sec:mixed} use the default sampling parameters of the engine.

\begin{table}[t]
\centering
\caption{Per-experiment serving configuration. GPU counts are per node, and ctx is the served context limit.}
\label{tab:protocol}
\resizebox{\linewidth}{!}{
\begin{tabular}{llll}
\toprule
\textbf{Experiment} & \textbf{Model} & \textbf{GPUs} & \textbf{ctx} \\
\midrule
Single-request, 4 engines & Qwen2.5-72B & A800$\times$4 & 131K \\
Single-request, 4 engines & Qwen2.5-7B-1M & A100$\times$1 & 131K \\
Single-request length sweep & Qwen2.5-72B & A800$\times$4, A800$\times$8 & 131K \\
Single-request length sweep & Qwen2.5-7B-1M & A100$\times$1, A800$\times$1 & 270K \\
Ultra-long 128K--1M & Qwen2.5-7B-1M & A800$\times$4 & 1.01M \\
Ultra-long, single GPU & Qwen2.5-7B-1M & A100$\times$1 & 1.01M \\
Concurrency sweep & Qwen2.5-7B-1M & A800$\times$4 & 40K \\
Mixed workload & Qwen2.5-72B & A800$\times$8 & 131K \\
Reasoning, single-request & Qwen3-235B-A22B & A800$\times$8 & 262K \\
Reasoning, single-request & gpt-oss-120b & A800$\times$4 & 131K \\
\bottomrule
\end{tabular}}
\end{table}

\noindent\textbf{KV-utilization profiling.} The kernel-level $\eta_{\mathrm{kv}}$ cited in Section~\ref{sec:estimator} was measured on one A100-80G (vLLM 0.11.0 with FlashAttention, SGLang 0.5.3 with FlashInfer, both torch 2.8.0, the same DCA-disabled 7B-1M checkpoint in bfloat16) by profiling 128 single-request decode steps per context length with the engine's built-in torch profiler, CUDA graphs disabled so decode kernels remain visible. Hardware byte counters were not available on this machine, so utilization is attention-kernel time measured against theoretical KV bytes, which assumes each KV byte is read once; the SGLang 131K point was measured twice end to end and reproduced within $0.4\%$.

\section{From Decode Latency to Serving Cost}
\label{sec:cost_from_latency}

The single-request scaling of Figure~\ref{fig:single-user-extrapolation} measures per-request
\emph{latency}, but a provider's unit economics is set by per-token \emph{cost}. Under the
per-second operating model of Section~\ref{sec:cost_model}, the cost of a generated token is
\begin{equation}
\$/\text{token} \;=\; \frac{G\cdot \$_{\text{GPU-sec}}}{\bar{q}} \;=\; \frac{\kappa_{\mathrm{node}}}{\bar{q}},
\label{eq:cost_per_token}
\end{equation}
where $\bar{q}$ is the server-side aggregate output throughput delivered by the $G$ GPUs of the
node. Equation~\ref{eq:cost_per_token} restates the break-even condition of
Section~\ref{sec:cost_model} as $\$/\text{token}\le r_{\mathrm{tok}}$. Because production serving is
batched, single-request latency gives an upper bound on cost, so we measure how
$\bar{q}$ grows with concurrency at different output lengths.

\noindent\textbf{Decomposing decode under concurrency.}
At decode step $t$ a batch of $B_t$ sequences issues \emph{one} read of the model weights $W$,
shared across all $B_t$ sequences, while \emph{each} sequence reads its own Key--Value cache. The
resulting decode cost separates into an amortizable weight term and a per-sequence KV term,
\begin{equation}
\text{Cost}_{\mathrm{dec}} \propto W\!\sum_t B_t^{-1}
+ k_{\mathrm{kv}}\big(L_{\mathrm{in}}L_{\mathrm{out}} + L_{\mathrm{out}}^2/2\big).
\label{eq:decode_decomp}
\end{equation}
The weight-read term corresponds to $c_w$, the part of the single-request slope $c = c_w + c_{\mathrm{ovh}}$ (Section~\ref{sec:decode_cost_model}) that shrinks with $G$; batching divides it by
$B$. The quadratic term is the $O(L_{\mathrm{out}}^2)$ traffic each sequence generates over its
own cache and, \emph{by construction}, cannot be shared across a batch. As $L_{\mathrm{out}}$ grows,
this un-amortizable term claims an increasing share of decode, so the throughput gain available from
batching must shrink.

\noindent\textbf{Measured concurrency scaling.}
Table~\ref{tab:batch_tput} confirms this trend. At short outputs ($2$K) throughput scales
near-linearly with concurrency (vLLM $42\times$ from $B{=}1$ to $B{=}64$), as expected when the
shared weight read dominates. At $32$K the same increase in concurrency yields $22.5\times$ on vLLM and $18.8\times$ on SGLang,
against an ideal $64\times$.

\label{sec:appendix_batch_tput}
\input{figs/tab_batch_tput}

\noindent\textbf{Batched form of the model.}
Equation~\ref{eq:decode_decomp} gives a per-step time $t_{\mathrm{step}} = c_{\mathrm{step}} + c_{\mathrm{seq}} B + 2d\sum_i \ell_i$, where $\ell_i$ is the context of sequence $i$ at that step; summed over a cell of $B$ equal-length requests it gives $T = c_{\mathrm{step}}\max_i L_i + c_{\mathrm{seq}}\sum_i L_i + d\sum_i L_i^2$. We fit the three coefficients by least squares on the vLLM grid of Table~\ref{tab:batch_tput}: concurrency $1$, $4$, $8$, $16$, $32$ and $64$ at $2$K, $8$K and $32$K. The fit gives $c_{\mathrm{step}}=5.69{\times}10^{-3}$\,s, $c_{\mathrm{seq}}=3.52{\times}10^{-5}$\,s and $d=4.10{\times}10^{-9}$\,s/tok$^2$, reproduces all $18$ cells within $2.8\%$ (mean $1.3\%$), and predicts two unseen cells of $65$K outputs at concurrency $4$ and $16$ within $1.0\%$ and $0.3\%$. We also test the form on batches of mixed lengths, which the equal-length grid cannot separate from a model that assigns every sequence the batch-average KV length. Six cells mix $2$ to $12$ requests of $32$K or $65$K with $2$K requests in batches of $16$ and $32$, and we run them on both engines. With the coefficients fitted on each engine's equal-length grid, the per-sequence form predicts all six within $5\%$ on SGLang and within $3$ to $5\%$ on vLLM 0.11.0, whereas the batch-average variant is off by $5$ to $13\%$ on SGLang and $8$ to $14\%$ on vLLM. On the cell that separates the two forms most, four $65$K and twelve $2$K requests on vLLM, the per-sequence form gives $459$\,s and the batch-average form $408$\,s, compared with $473$\,s measured.

\noindent\textbf{Transfer across hardware.}
Extra devices speed up a single request sub-linearly. On Qwen2.5-14B, $c$ falls by $1.58\times$, $2.33\times$ and $2.66\times$ at $G{=}2$, $4$ and $8$ relative to one GPU. We separate the device count from the GPU type with two controlled comparisons on Qwen2.5-7B-1M, vLLM 0.11.0 and repetition penalty $1.0$, each a sweep over six output lengths from $2$K to $262$K. From one to four A800s, $d$ falls from $2.01{\times}10^{-8}$ to $5.65{\times}10^{-9}$\,s/tok$^2$, a factor of $3.55$ compared with the ideal $4$, whereas $c$ falls from $1.12{\times}10^{-2}$ to $5.70{\times}10^{-3}$\,s/tok, a factor of $1.97$. The two device counts fix $c = a_c/G + b_c$ at $a_c = 7.4{\times}10^{-3}$ and $b_c = 3.9{\times}10^{-3}$\,s/tok. The weight read shrinks with $G$ and a fixed per-step overhead of about $3.9$\,ms does not. The break length follows the two-term form of Section~\ref{sec:cost_model}. Its model-and-engine part $c_w/d$ is $368$K on one A800 and $326$K on four; its overhead part $c_{\mathrm{ovh}}/d$ is $192$K and $683$K, so $L^\star$ shifts from $560$K to $1.01$M when the device count grows. The A100 and the A800 share one GPU architecture and the same HBM bandwidth. With one card each, $c$ differs by $2\%$ and $d$ by $8\%$, and the A800 coefficients predict the A100 decode time within $1.5\%$ at $262$K. The $\$/\text{token}$ estimate of Equation~\ref{eq:cost_per_token} extends to new accelerators from published GPU-second prices once $c$ and $d$ are rescaled this way, in line with the GPU-sec/token $\times$ \$/GPU-h costing of \citet{erdil2025}.

\section{Mixed-Workload Latency Sweeps}
\label{sec:appendix_mixed_sweeps}

\begin{figure}[h!]
\centering
\includegraphics[width=0.99\columnwidth]{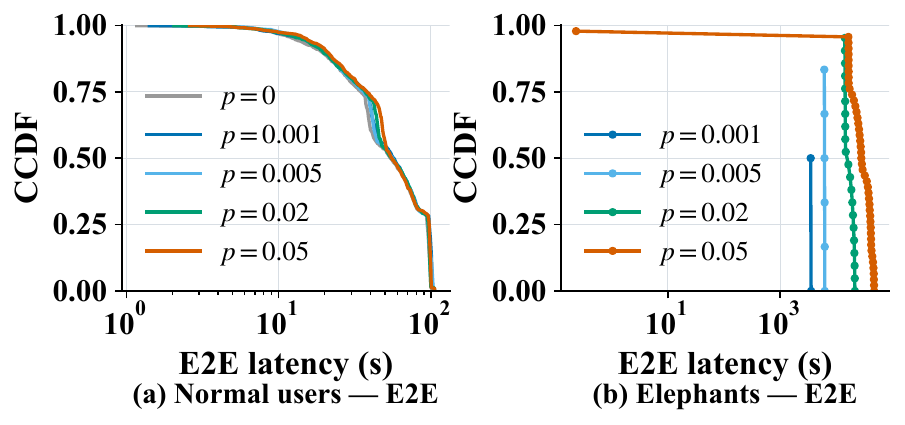}
\caption{Mixed-workload elephant-fraction latency sweep on Qwen2.5-72B-Instruct (vLLM, $\eta=1.0$, $N=1000$).}
\label{fig:ablation_plong_main}
\end{figure}

Figure~\ref{fig:ablation_plong_main} shows per-class latency distributions as the elephant fraction $p_{\mathrm{ele}}$ rises from $0$ to $0.05$ under the burst launch of Section~\ref{sec:mixed}. The normal-class curves stay in a narrow range across the sweep, whereas the elephant-class curves span one to three orders of magnitude as more elephants share the cache; Section~\ref{sec:ablation_main} reports the cache-capacity transition behind that spread.

\section{Prediction at Unseen Lengths on Hybrid-Attention Models}
\label{sec:appendix_hybrid}

Table~\ref{tab:hybrid_heldout} applies the two-point protocol of Table~\ref{tab:heldout} to two models whose attention is not uniform across layers, with $k_{\mathrm{kv}}$ counting only the layers whose cache grows with length (Appendix~\ref{sec:cost_model_derivation}). gpt-oss-120b runs on vLLM 0.11.0 with TP$=4$ on A800s; Qwen3.5-397B-A17B-FP8 runs on vLLM 0.19.1 with TP$=8$ on A800s, with ten length points from $1$K to $131$K. The calibration points are the two shortest elephant lengths of each sweep, where the quadratic term already exceeds the per-step jitter.

\begin{table}[h]
\centering
\caption{Decode-time prediction at unseen lengths on two hybrid-attention models, calibrated on the two points in parentheses. Error is (predicted $-$ measured)/measured.}
\label{tab:hybrid_heldout}
\resizebox{0.8\linewidth}{!}{
\begin{tabular}{llrr}
\toprule
\textbf{Model (calibration)} & $L_{\mathrm{out}}$ & \textbf{Measured} & \textbf{Error} \\
\midrule
\multirow{3}{*}{\shortstack[l]{gpt-oss-120b\\(13K, 39K)}}
 & 66K  & 2\,034\,s & $-1.3\%$ \\
 & 92K  & 3\,048\,s & $-1.6\%$ \\
 & 118K & 4\,178\,s & $-1.9\%$ \\
\midrule
\multirow{3}{*}{\shortstack[l]{Qwen3.5-397B-A17B\\(32K, 64K)}}
 & 96K  & 1\,178\,s & $-0.4\%$ \\
 & 118K & 1\,469\,s & $-0.9\%$ \\
 & 131K & 1\,640\,s & $-0.9\%$ \\
\bottomrule
\end{tabular}}
\end{table}

\section{Shared Serving with LMSYS-Chat-1M Prompts}
\label{sec:appendix_lmsys_cell}

All shared-serving cells of Section~\ref{sec:mixed} draw normal prompts from WildChat-1M. As a robustness check, we repeat the Poisson cell of Table~\ref{tab:slo_single} ($p_{\mathrm{ele}}{=}0.01$, $1$\,req/s for $900$\,s, Qwen2.5-72B, TP$=8$) with normal prompts drawn from LMSYS-Chat-1M, on vLLM 0.19.1; the reference WildChat cell ran on vLLM 0.11.0, so the comparison spans both the dataset and the engine version. The cell draws the same $892$ normal requests and $9$ elephants. The median normal TTFT is $0.12$\,s in both cells, and $99.4\%$ of normal requests meet both bounds versus $100\%$; the five misses arrive in the first $25$\,s of the run, before the first elephant, and include the server's start-up delay. The queue stays empty and no request is preempted in either cell. LMSYS-Chat-1M prompts are shorter ($58$ versus $154$ tokens on average) and its responses stop earlier, so the median normal E2E falls from $11.7$ to $5.4$\,s; the elephant requests behave in the same way.

\section{Examples of Ultra-Long Responses in the Wild}
\label{sec:appendix_examples}

Table~\ref{tab:long_response_examples} summarizes representative cases of ultra-long LLM outputs arising from diverse scenarios; their natural occurrence is what motivates the elephant request as a proxy.

\begin{table*}[t]
\centering
\caption{Representative real-world sources of ultra-long LLM outputs. Reasoning and writing tasks routinely produce responses of 5K–30K+ tokens. Example prompts and response descriptions are illustrative.}
\vspace{-1ex}
\label{tab:long_response_examples}
\resizebox{\textwidth}{!}{
\begin{tabular}{llp{6cm}p{6cm}}
\toprule
\textbf{Source} & \textbf{Typical Length} & \textbf{Example Prompt} & \textbf{Response Content} \\
\midrule
Reasoning model (DeepSeek-R1~\cite{deepseek_r1}) & 10K–30K tokens & ``Prove that $\sqrt{2}$ is irrational.'' & Extended chain-of-thought reasoning: \texttt{<think>Let me consider... actually, let me try a different approach... [... reasoning omitted ...]</think>} followed by a concise final answer. \\
\addlinespace
Long-form writing (LongWriter~\cite{longwriter}) & 2K--32K words & ``Write a 10,000-word essay on the history of climate science.'' & Structured multi-section document with introduction, body sections, and conclusion; output length determined by the task specification, not the prompt length. \\
\addlinespace
Induced long-output case (Engorgio~\cite{engorgio}) & Up to $\sim$4K tokens & Prompt crafted to induce non-terminating generation. & Repetitive self-reinforcing continuations; output extends far beyond the natural stopping point. Unlike other cases, content quality degrades progressively. \\
\bottomrule
\end{tabular}}
\vspace{-1ex}
\end{table*}

\section{Length Distribution of Real Workloads}
\label{sec:appendix_workload_length}

To characterize the natural workload, we tokenize both prompts and assistant responses across nine publicly available datasets spanning ordinary chat (LMSYS-Chat-1M~\cite{lmsys_chat_1m}, OASST1~\cite{oasst1}, ShareGPT-52K~\cite{sharegpt52k}, WildChat-1M~\cite{wildchat}), long-form generation (LongWriter-6K~\cite{longwriter}), and DeepSeek-R1-style reasoning (Dolphin-R1-DeepSeek~\cite{dolphin_r1}, OpenR1-Math-220K~\cite{openr1_math}, Bespoke-Stratos-17K~\cite{bespoke_stratos}, OpenThoughts-114K~\cite{openthoughts}). Tokenization uses the \texttt{DeepSeek-R1-Distill-Qwen-1.5B} tokenizer with \texttt{add\_special\_tokens=False}. Table~\ref{tab:workload_length} reports per-dataset output (decode) length percentiles. The picture is consistent across sources: most traffic is short. Chat responses have $P_{99}$ below 4K tokens (OASST1 861, ShareGPT-52K 2\,464, WildChat-1M 1\,561). Long-form and reasoning are the only sources that push outputs meaningfully longer, and even they concentrate around a $P_{99}$ of 15--22K tokens (LongWriter-6K $\sim$15K; the four reasoning sets 16--22K), with maxima reaching only $\sim$34--88K. These lengths mark where super-linear growth begins, and current APIs already allow single responses of 128K tokens or more (Section~\ref{sec:experiments}).

\begin{table}[h]
\centering
\caption{Decode-length percentiles across natural workload sources, computed at the tokenizer level. \emph{Decode tokens} = assistant response (chat / long-form) or reasoning chain plus visible final answer (reasoning). $n$ counts parseable rows in the first 10\,000 sampled.}
\label{tab:workload_length}
\resizebox{\linewidth}{!}{
\begin{tabular}{llrrrrrrr}
\toprule
\textbf{Dataset} & \textbf{Kind} & $n$ & $P_{50}$ & $P_{90}$ & $P_{95}$ & $P_{99}$ & $P_{99.9}$ & max \\
\midrule
LMSYS-Chat-1M       & chat       & 10\,000 &  130 &   394 &   441 &   588 &     993 & 109\,052 \\
OASST1              & chat       &  6\,722 &  149 &   454 &   561 &   861 &  1\,703 &  2\,913 \\
ShareGPT-52K        & chat       & 10\,000 &  292 &   922 & 1\,217 & 2\,464 &  3\,945 & 60\,276 \\
WildChat-1M         & chat       &  9\,833 &  120 &   771 & 1\,009 & 1\,561 &  3\,409 &  5\,693 \\
LongWriter-6K       & long-form  &  5\,995 &  232 & 7\,573 & 9\,406 & 15\,036 & 27\,200 & 34\,056 \\
\midrule
Dolphin-R1-DeepSeek & reasoning  & 10\,000 & 1\,838 & 7\,639 & 10\,786 & 17\,373 & 24\,240 & 41\,746 \\
OpenR1-Math-220K    & reasoning  & 10\,000 & 4\,282 & 10\,568 & 12\,713 & 15\,715 & 17\,808 & 20\,574 \\
Bespoke-Stratos-17K & reasoning  & 10\,000 & 2\,258 & 6\,842 & 9\,706 & 16\,278 & 21\,711 & 33\,290 \\
OpenThoughts-114K   & reasoning  & 10\,000 & 5\,516 & 15\,583 & 18\,218 & 22\,398 & 27\,753 & 88\,030 \\
\bottomrule
\end{tabular}}
\end{table}

\section{Continuous Arrival with Resident Elephants}
\label{sec:appendix_arrival_ablation}

The main-body mixed-workload cells use a burst launch (all requests issued at $t{=}0$), which queues most requests at once; Table~\ref{tab:slo_single} already separates that queueing from elephant decode by re-running the same configuration under Poisson arrival. This appendix measures normal-request throughput while the number of running elephants grows under continuous arrival.

\begin{figure}[h]
\centering
\includegraphics[width=0.95\columnwidth]{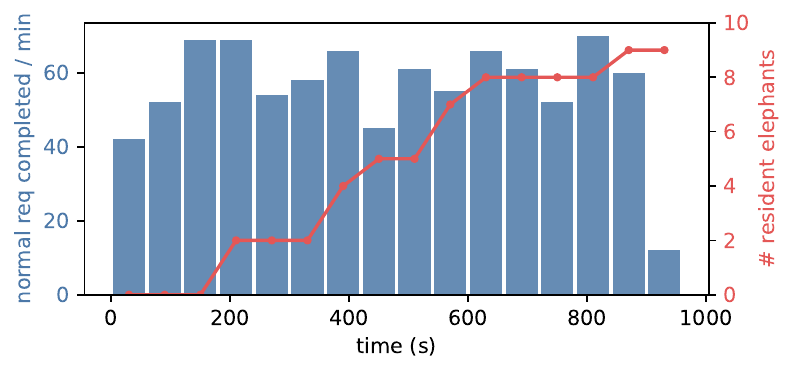}
\caption{Steady-state cell: normal requests completed per minute (bars) stay flat as the number of resident ultra-long elephant requests (line) grows from $0$ to $9$; the final dip is where the $900$\,s arrival window ends. Continuous arrival keeps normal throughput up despite resident elephants.}
\label{fig:steady_tput}
\end{figure}

In a lightly loaded regime ($p_{\mathrm{ele}}{=}0.01$, $\eta{=}0.3$, Poisson normal traffic at $1$\,req/s for $900$\,s, aggregate KV below capacity), normal requests keep completing at the arrival rate while the resident elephant population grows from $0$ to $9$ (Figure~\ref{fig:steady_tput}), and the median normal TTFT stays at $0.15$\,s. The near-idle tail of the burst traces therefore comes from the end of arrivals, after which only elephants remain. In this cell the elephants are a persistent minority that hold KV cache and running slots, which is an opportunity cost for the provider.

\section{Ablation Detail: Per-Model CCDFs and Hardware Comparison}
\label{sec:appendix_ablation_detail}

This appendix extends Section~\ref{sec:ablation_main} with detailed per-class percentile tables, per-model CCDFs across the elephant-fraction sweep, and a hardware comparison (A800 vs.\ H800).

\subsection{SLO and Concurrency Summary Tables}
\label{sec:appendix_slo_concurrency_tables}

\noindent Table~\ref{tab:concurrency_summary} lists peak and mean concurrency for the cells of Figure~\ref{fig:kv}, together with the wall time each cell took. The two concurrency numbers differ widely. The node briefly runs several hundred requests at once, then averages a few dozen across the cell, because most of the wall time is the elephant-only phase in which only the elephants are running. The waiting counts show where requests queue. The 72B cell carries a backlog of $126$ requests on average, while the gpt-oss-120b cell carries almost none and still runs more than four times as long, its elephants set to the full context window rather than $40\%$ of it.

\begin{table}[h]
\centering
\resizebox{\linewidth}{!}{\input{figs/ablation/concurrency_table.tex}}
\caption{Peak and mean numbers of running and waiting requests, and cell wall time, for the cells in Figure~\ref{fig:kv}.}
\label{tab:concurrency_summary}
\end{table}

\subsection{Per-Model $p_{\mathrm{ele}}$ CCDFs}
\label{sec:appendix_perclass_plong}

\noindent The same split holds on the dense 14B chat model and the two reasoning models. Their normal-user CCDFs cluster tightly across $p_{\mathrm{ele}}\in\{0, 0.01, 0.02, 0.05, 0.10\}$, whereas elephant CCDFs span one to three orders of magnitude. The reasoning models (235B, 120b) carry larger absolute elephant latencies because each generation contains a long thinking phase, but the normal-vs-elephant gap has the same shape as on the chat models.

\subsection{Hardware Comparison: Qwen3.5-397B-A17B-FP8 on A800 vs.\ H800}
\label{sec:appendix_hardware}

\begin{figure}[!htb]
\centering
\includegraphics[width=0.85\columnwidth]{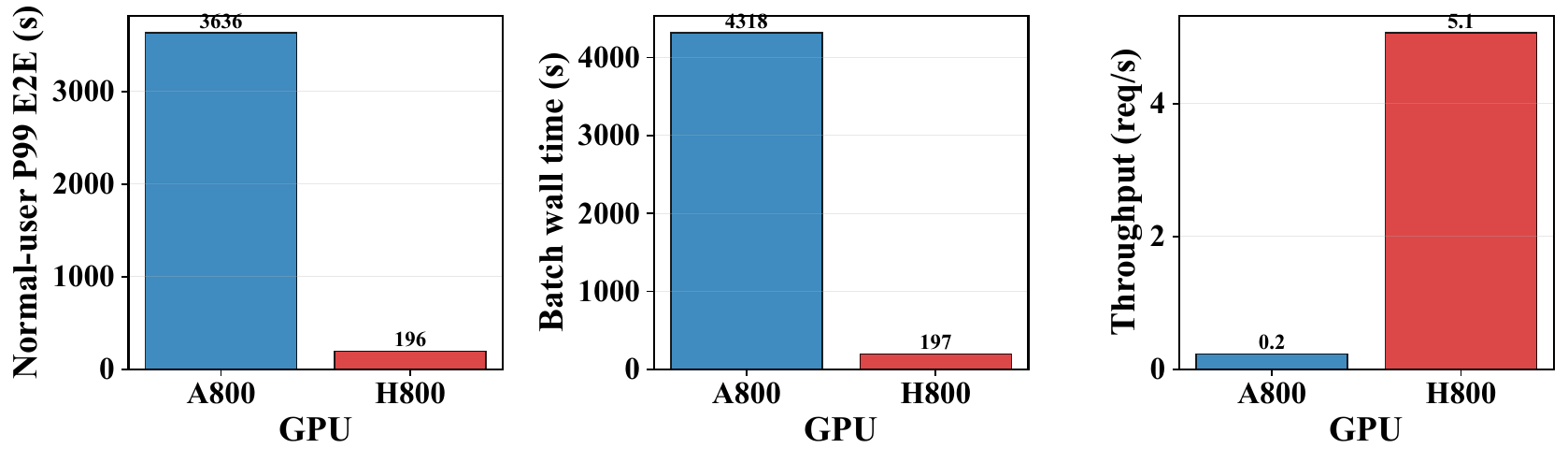}
\caption{Hardware comparison of the Qwen3.5-397B-A17B-FP8 cell ($p_{\mathrm{ele}}=0.01$, $\eta=0.5$, $N=1000$, vLLM TP$=$8) on A800 and H800. The higher-bandwidth H800 improves every metric, reducing normal-user $P_{99}$ E2E by about an order of magnitude. The two cells drew different elephant counts, so the comparison indicates only the direction of the effect.}
\label{fig:appendix_hardware}
\end{figure}

\noindent We run the Qwen3.5-397B-A17B-FP8 cell on an H800 node with the same vLLM TP$=$8 configuration (Figure~\ref{fig:appendix_hardware}). Higher memory bandwidth improves every metric in Figure~\ref{fig:appendix_hardware}: the H800 cell completes the workload faster and lowers normal-user $P_{99}$ E2E by about an order of magnitude, and the coefficient fits of Section~\ref{sec:estimator} quantify that dependence.

%% file: figs/tab_batch_tput.tex
\begin{table}[t]
\centering
\caption{Decode throughput vs.\ concurrency. Server-side output tokens/sec for
Qwen2.5-7B-Instruct-1M ($4\times$A800, TP$=4$, full attention) on vLLM and SGLang with matched sampling
(repetition penalty $1.0$), sweeping the number of concurrent decode sequences $B$ at three fixed output lengths
$L_{\mathrm{out}}$. Both engines use continuous batching; $B$ is the offered concurrency, not a static batch size.
The final column is the $B{=}1\!\to\!B{=}64$ throughput gain (ideal linear scaling would be $64\times$).
Short outputs scale near-linearly; at $32$K the gain falls to $22.5\times$ (vLLM) and $18.8\times$ (SGLang).
The SGLang $2$K single-stream cell is omitted because that measurement carries first-request warmup.}
\label{tab:batch_tput}
\resizebox{\linewidth}{!}{
\begin{tabular}{llrrrr}
\toprule
\textbf{Engine} & $\mathbf{L_{\mathrm{out}}}$ & $\mathbf{B{=}1}$ & $\mathbf{B{=}16}$ & $\mathbf{B{=}64}$ & \textbf{$\times$($\tfrac{B64}{B1}$)} \\
\midrule
\multirow{3}{*}{vLLM}
 & 2K  & 175.3 & 2532.3 & 7380.2 & 42.1$\times$ \\
 & 8K  & 175.1 & 2401.8 & 6438.4 & 36.8$\times$ \\
 & 32K & 171.1 & 1923.9 & 3841.8 & 22.5$\times$ \\
\midrule
\multirow{3}{*}{SGLang}
 & 2K  & --- & 2614.9 & 7493.7 & --- \\
 & 8K  & 218.2 & 2463.9 & 6433.4 & 29.5$\times$ \\
 & 32K & 207.6 & 2001.2 & 3893.8 & 18.8$\times$ \\
\bottomrule
\end{tabular}}
\end{table}

%% file: figs/ablation/concurrency_table.tex
\begin{tabular}{llrrrrr}
\toprule
Model & Cell & Peak running & Mean running & Peak waiting & Mean waiting & Cell wall (s) \\
\midrule
Qwen2.5-72B & p=0.010, $\eta$=0.40 & 521 & 35.5 & 254 & 126.0 & 14432 \\
gpt-oss-120b & p=0.010, $\eta$=1.00 & 510 & 18.6 & 8 & 4.9 & 66480 \\
\bottomrule
\end{tabular}